\pdfoutput=1

\documentclass[11pt]{article}

\usepackage[]{emnlp2023}

\usepackage[export]{adjustbox}
\usepackage{float}
\usepackage{graphicx}
\usepackage{makecell}
\usepackage{enumitem}
\usepackage{arydshln}
\usepackage{booktabs}
\usepackage{tabularx}
\usepackage{multirow}
\usepackage{xcolor}
\usepackage{bbding}
\usepackage{amsmath}
\usepackage{amsfonts}
\usepackage{pgfplots}
\pgfplotsset{compat=1.7}
\usepackage{tabu}
\usepackage{subfigure}
\usepackage{pifont}
\usepackage{colortbl}

\usepackage{CJK}
\usepackage{algorithm}
\usepackage{fdsymbol}
\usepackage{algpseudocode}

\usepackage{times}
\usepackage{latexsym}

\usepackage{fdsymbol}
\usepackage{fontawesome}

\definecolor{ForestGreen}{RGB}{34,139,34}
\definecolor{BrickRed}{rgb}{.72,0,0}
\definecolor{LakeBlue}{RGB}{0,61,153}
\newcommand{\fstar}{\textsuperscript{\fontsize{6pt}{6pt}\selectfont \faStarO}}
\newcommand{\fmoon}{\textsuperscript{\fontsize{6pt}{6pt}\selectfont \faMoonO}}

\usepackage[T1]{fontenc}

\usepackage[utf8]{inputenc}

\usepackage{microtype}

\usepackage{inconsolata}

\title{Make Prompt-based Black-Box Tuning Colorful: Boosting Model Generalization from Three Orthogonal Perspectives}

\author{
Qiushi Sun\textsuperscript{$\heartsuit$}\quad \quad
Chengcheng Han\textsuperscript{$\diamondsuit$}\quad \quad
Nuo Chen\textsuperscript{$\diamondsuit$}\\
\bf{
Renyu Zhu\fstar \quad \quad
Jingyang Gong\fmoon \quad \quad
Xiang Li\textsuperscript{$\diamondsuit$}\thanks{~~Corresponding author.} \quad \quad
Ming Gao\textsuperscript{$\diamondsuit$}
}\\
\textsuperscript{$\heartsuit$}National University of Singapore
\textsuperscript{$\diamondsuit$}East China Normal University\\
\fstar NetEase Fuxi AI Lab \fmoon New York University \\
\texttt{qiushisun@u.nus.edu}, \texttt{\{chengchenghan,nuochen\}@stu.ecnu.edu.cn} \\
\texttt{zhurenyu@corp.netease.com}, \texttt{jingyang.gong@nyu.edu} \\ \texttt{\{xiangli,mgao\}@dase.ecnu.edu.cn}
}

\begin{document}



\maketitle

\begin{abstract}
Large language models (LLMs) have shown increasing power on various natural language processing (NLP) tasks.
However, 
tuning these models for downstream tasks usually 
needs
exorbitant costs or is unavailable due to commercial considerations.
Recently, 
black-box tuning has been proposed to address this problem by optimizing task-specific prompts without accessing the gradients and hidden representations.
However, 
most existing works have
yet 
fully exploited the potential of gradient-free optimization under the scenario of few-shot learning.
In this paper, 
we describe BBT-RGB, 
a suite of straightforward and complementary techniques for enhancing the efficiency and performance of black-box optimization. 
Specifically,
our method includes three plug-and-play components:
(1) Two-stage derivative-free optimization strategy that facilitates fast convergence and mitigates overfitting; 
(2) Automatic verbalizer construction with its novel usage under few-shot settings;
(3) Better prompt initialization policy based on instruction search and auto-selected demonstration.
Extensive experiments across various tasks 
on
natural language understanding and inference demonstrate the effectiveness of our method. 
Our codes and data are available at \url{https://github.com/QiushiSun/BBT-RGB}.

\end{abstract}

\section{Introduction}
\label{sec:intro}
Transformer-based Language models~\cite{vaswani2017attention} have achieved remarkable improvements among various NLP tasks~\cite{qiu2020pre, lin2022transformer} in recent years.
These models are mainly first pre-trained on a large-scale unsupervised corpus and then fine-tuned on a specific downstream task.
However, 
this paradigm of {pre-train and fine-tune} face challenges in the era of Large Language Models (LLMs)~\cite[][\emph{inter alia}]{brown2020GPT3, ouyang2022trainingLM, chowdhery2022palm, zhang2022opt, scao2022bloom, touvron2023llama, openai2023gpt4}.
The ever-growing model size leads to a non-stop increase in the cost of tuning, and deploying separate copies of LLMs in real applications becomes exorbitantly expensive.
Though recent research on parameter-efficient tuning~\cite[][\emph{inter alia}]{li2021prefix, lester2021power} alleviates the problem by tuning a small percentage of parameters while keeping the backbone frozen, 
the second problem arises: \textit{most LLMs are released as a service, and users can only access them through Black-Box APIs.} 
This implies that the aforementioned tuning strategies become less viable owing to the inaccessibility of parameters and gradients, 
thereby causing a dilemma for downstream applications.
\citet{sun2022bbt} describe this scenario as Language Model-as-a-Service (LMaaS): 
Users are unable to tune the model parameters but can accomplish the tasks of interest by finding appropriate prompts with limited examples. 
Then, Black-Box Tuning (BBT) is proposed as a framework for derivative-free optimization under few-shot settings. 
Recently, BBTv2~\citep{sun2022bbtv2}  has been presented as an improved version that prepends prompts to hidden states of models instead of only injecting prompt tokens in the input layer.
However, 
the potential of black-box optimization is still not fully exploited.
Previous tuning methods are prone to overfit / fall into local optimum under the scenario of few-shot learning. 
This phenomenon is triggered by both the characteristics of the Derivative-free optimization (DFO) algorithm and the unavailability of pre-trained prompts under few-shot settings. 


In this paper, 
we present BBT-RGB, 
a suite of straightforward, 
complementary, and pluggable techniques that further explore the possibility of black-box tuning. 
We take one step forward in black-box tuning from the following three aspects 
1) Employing a two-stage DFO strategy for the attenuation of overfitting.
2) Utilizing multiple auto-selected verbalizers to exploit the context further.
3) Combining manual prompt with new search approach for task instructions improvement. 

Extensive experiments across various NLP downstream tasks demonstrate the superiority of our method.
Besides,
BBT-RGB can significantly outperform current gradient-based Parameter-Efficient tuning methods~\cite{houlsby2019ParamEff, Zaken2022bitfit, hu2022lora, liu2022ptuning} under the scenario of few-shot learning.

Our main contributions can be summarized as follows:
\begin{itemize}[itemsep=6pt,topsep=0pt,parsep=0pt]
    \item We propose a two-stage derivative-free optimization strategy that enables stable convergence of training tunable prompts while effectively mitigating the issue of overfitting.
    \item To further exploit the LLM's output, we propose a verbalizer selection process to derive multiple appropriate candidates. Moreover, 
    instruction with judiciously selected demonstration is adopted for prompt initialization.
    \item 
    A wide range of NLP tasks is covered to verify the effectiveness of our approach. 
    By employing our method, 
    optimization\footnote{
    We follow bbtv2~\cite{sun2022bbtv2} to use random projection matrices to transform prompt parameters into low-dimensional subspaces.
} under the derivative-free framework can reach comparative performance to full fine-tuning.
\end{itemize}

\begin{figure*}[ht!]
    \centering
    \includegraphics[width=0.88\linewidth]{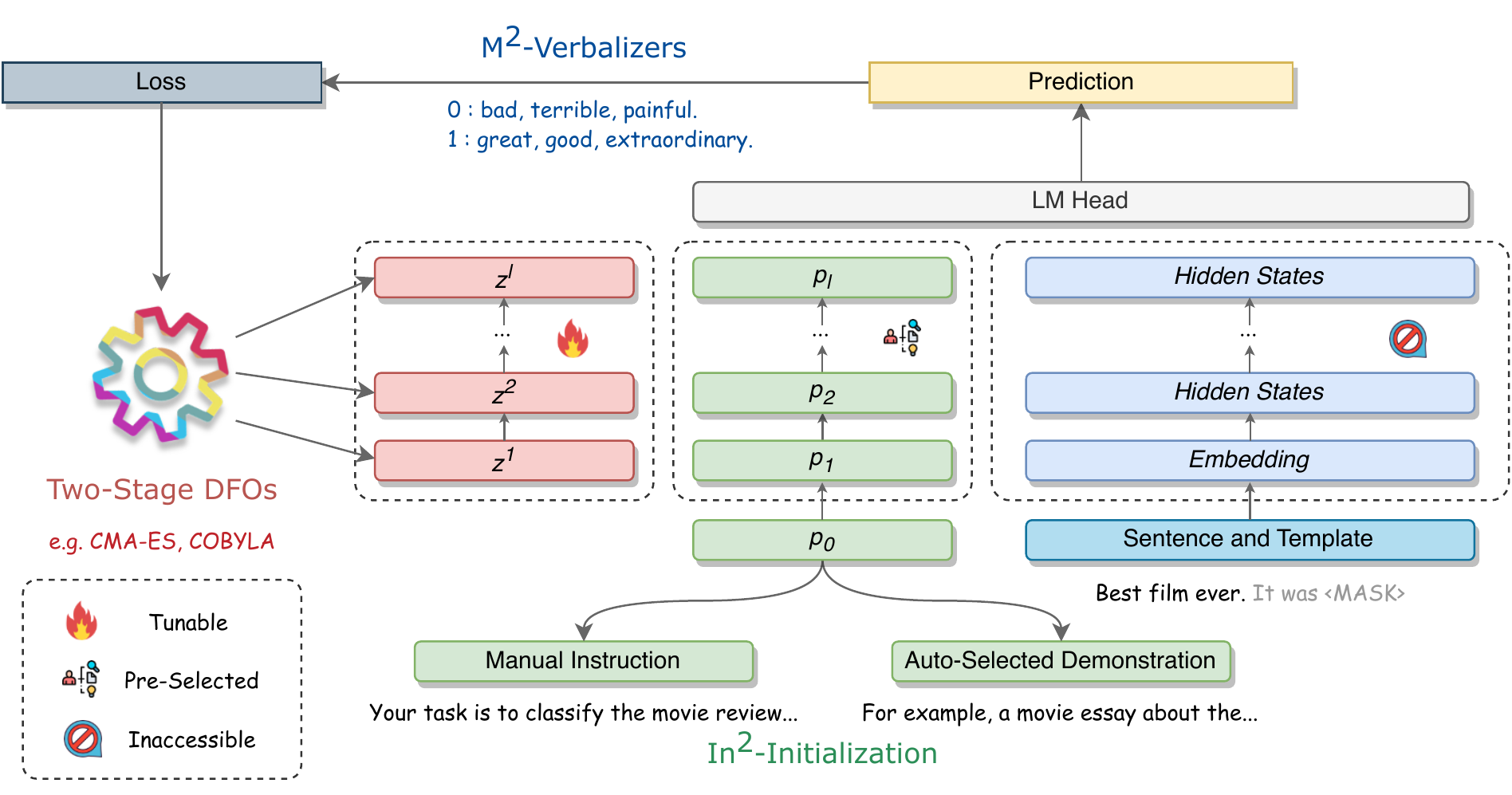}
    \caption{An illustration of BBT-RGB. 
    Given a backbone model with $L$ layers.
    The target is to optimize continuous prompts $z^{l}, l\in [1, L]$.
    We use {\color{BrickRed}{\textbf{R}ed}}, {\color{ForestGreen}{\textbf{G}reen}} and {\color{LakeBlue}{\textbf{B}lue}} to indicate three distinct aspects of our strategy, which inspired the naming of our method. 
    {\color{LakeBlue}{M\textsuperscript{2}} Verbalizers} (Multi-Mixed Verbalizers) further utilize the information provided by the LLMs.
    {\color{ForestGreen}{In\textsuperscript{2}} Initialization} (Instruction learning + In-context learning) improves prompt-based tuning by integrating both instruction and demonstration,
    noted as $p_{l }$.
    And {\color{BrickRed}{Two-Stage DFOs}} exploit the advantages of different optimization methods. \includegraphics[width=.35cm, valign=c]{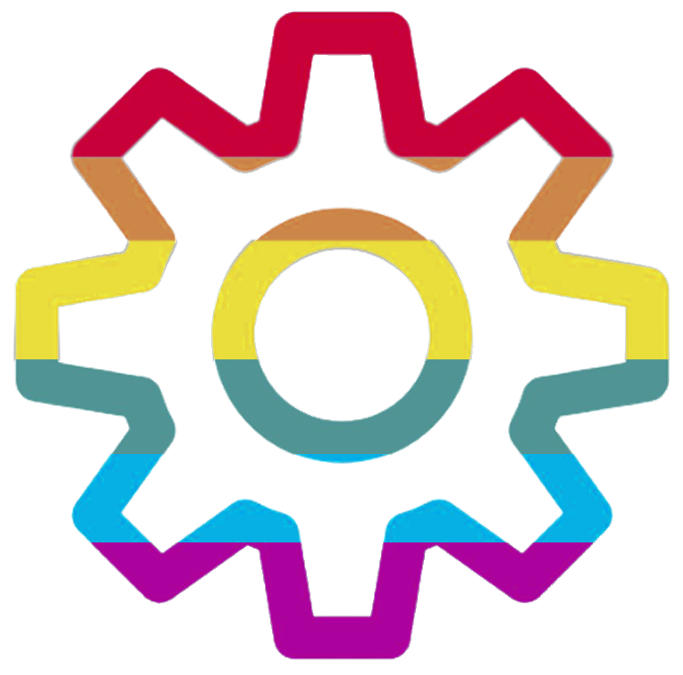}\ represents the combination of derivative-free optimizers.
    (Best viewed in color.)
    }
    \label{fig:overview}
    \vspace{-0.75em}
\end{figure*}

\section{Preliminaries}


 \subsection{Large Language Models and APIs}

Large language models (LLMs)~\cite{devlin2018bert, liu2019roberta, brown2020GPT3} have revolutionized the NLP landscape in the past few years. 
Given some examples of tasks as input, 
LLMs can be ``prompted'' to conduct a wide range of NLP tasks.
These huge models are usually released as a service~\cite{brown2020GPT3, chen2021evaluating, ouyang2022trainingLM}, which allows users to interact with the models deployed on the cloud servers through APIs.
Unlike some popular open-source LMs~\cite{devlin2018bert, liu2019roberta} that can be directly utilized by researchers, 
access to the parameters and gradients of LLMs is restricted due to commercial, ethical, and security concerns.

\subsection{Prompt-based Learning}

Prompt-based learning~\cite{liu2021promptsurvey} transforms an NLP downstream task into a masked language modeling (MLM) task and narrows the discrepancy between pre-training and fine-tuning.
Based on the prompt format, 
prompt-based learning can be categorized into discrete prompts and continuous prompts.
Discrete prompts can be designed manually~\citep{brown2020GPT3, schick2020automatically} or generated automatically~\cite{gao2021making}.
Continuous prompts are designed as a sequence of vectors~\cite{qin2021learning, lester2021power} that are usually prepended to the input and optimized by gradients.
Recently,~\citet{sun2022bbt} propose BBT for optimizing prompts under gradient-free settings, as is shown in section~\ref{sec:bbt}.
We mainly focus on the optimization of continuous prompts under the black-box settings in this paper.



\subsection{Derivative-free Optimization}

Derivative-free optimization (DFO) algorithms are capable of solving complex problems without the back-propagation process. 
DFO generally employs a sampling-and-updating framework~\cite{rios2013DFOreview, wierstra2014nes, qian2016dfo} to improve the solution iteratively.
For instance, 
Covariance Matrix Adaptation Evolution Strategy~\cite{hansen2001cma, hansen2003reducing}, namely CMA-ES, 
is a widely adopted evolutionary algorithm for non-linear non-convex continuous optimization.
At each iteration, 
the algorithm samples new potential solutions from a parameterized distribution model (\textit{e.g.}, multivariate normal distribution).
Besides,
we have COBYLA algorithm (Constrained Optimization BY Linear Approximation)~\cite{powell1994direct, powell1998direct} that builds a linear approximation model of the objective function and constraints within a trust region, 
iteratively updating the model based on the progress made in minimizing the objective function.


\section{BBT-RGB}
As is shown in Figure~\ref{fig:overview},
we introduce our method: BBT-RGB, 
which contains three orthogonal optimization perspectives of derivative-free learning\footnote{The backgrounds and formal definition of derivative-free learning methods are given in section~\ref{sec:bbt}.}.

\subsection{Two-Stage DFOs}
Previous works of black-box tuning mainly use CMA-ES to optimize the intrinsic dimensionality~\cite{aghajanyan2021intrinsic} of LLMs.
Nonetheless,
in the early training stage, 
the evolutionary algorithm (EA) exhibits a considerably faster convergence rate compared to the search-based algorithm (SA),
which potentially causes fast overfitting.
Then, the following steps would be futile.
Thus, we design a novel two-stage DFO algorithm\footnote{Due to space limitations, we put the detailed algorithm in Appendix~\ref{sec:two-stage-detailed}.
Algorithm~\ref{alg:two-stage} stands for a simplified version.} for black-box tuning, 
as is shown in algorithm~\ref{alg:two-stage}. 

\begin{algorithm}[t]
	\caption{Two-Stage DFOs}
	\label{alg:two-stage}
	\begin{algorithmic}[1]
		\Require \
		 popsize:$\lambda$, intrinsic dimension:$d$
            \Require \
          budget1:$b1$, budget2:$b2$, backbone:$f_{model}$
		\Ensure \
		hidden variable: $z$
		\Function{Two-Stage DFO}{}   
            \Repeat
            \For{each hidden layer}
            \State Update $z$ by Evolutionary DFO
            \EndFor
            \Until{ $b1$ times $f_{model}$ call}
            \For{each hidden layer}
            \Repeat
            \State Update $z$ by Search-based DFO
            \Until{ $b2//d$ times $f_{model}$ call}
            \EndFor
	    \EndFunction
	\end{algorithmic}
\end{algorithm}

We leverage the advantages of two different kinds of DFOs respectively.
In stage \uppercase\expandafter{\romannumeral1}, we use EA to perform coarse-grained population-level optimization, which has a specific budget (Number of API Calls) to move toward the target swiftly.
And the SA will use the remaining budgets in stage \uppercase\expandafter{\romannumeral2} for approximating the solution by dimension-level fine-grained search.

\begin{table*}[t]
\centering
\label{tab:main_results}
\resizebox{1.01\linewidth}{!}{
\begin{tabular}{lcccccccr}
\toprule
\multirow{2}{*}{\textbf{Method}} & \textbf{SST-2} & \textbf{Yelp P.} & \textbf{AG's News} & \textbf{DBPedia} & \textbf{MRPC} & \textbf{SNLI} & \textbf{RTE} & \multirow{2}{*}{\textbf{Avg.}} \\
& acc  & acc  & acc & acc & F1 & acc & acc & \\ 
\midrule
\multicolumn{9}{c}{\textit{Gradient-Based Methods}} \\ 
\midrule
Model Fine-Tuning & {85.39} \small{$\pm$2.84} & {91.82} \small{$\pm$0.79} & 86.36 \small{$\pm$1.85} & {97.98} \small{$\pm$0.14} & {\textbf{77.35}} \small{$\pm$5.70} & 54.64 \small{$\pm$5.29} & {\textbf{58.60}} \small{$\pm$6.21} & {78.88}\\
Prompt Tuning~\cite{lester2021power} & 68.23 \small{$\pm$3.78} & 61.02 \small{$\pm$6.65} & 84.81 \small{$\pm$0.66} & 87.75 \small{$\pm$1.48} & 51.61 \small{$\pm$8.67} & 36.13 \small{$\pm$1.51} & {54.69} \small{$\pm$3.79} & 63.46 \\
P-Tuning v2~\cite{liu2022ptuning} & 64.33 \small{$\pm$3.05} & {\textbf{92.63}} \small{$\pm$1.39} &83.46 \small{$\pm$1.01} &97.05 \small{$\pm$0.41} &68.14 \small{$\pm$3.89} &36.89 \small{$\pm$0.79} &50.78 \small{$\pm$2.28} & 70.47 \\ 
Adapter~\cite{houlsby2019ParamEff} & 83.91 \small{$\pm$2.90} & 90.99 \small{$\pm$2.86}  & 86.01 \small{$\pm$2.18} & {\textbf{97.99}} \small{$\pm$0.07} & 69.20 \small{$\pm$3.58} & {57.46} \small{$\pm$6.63} & 48.62 \small{$\pm$4.74} & 76.31 \\ 
LoRA~\cite{hu2022lora} & {\textbf{88.49}} \small{$\pm$2.90} & 90.21 \small{$\pm$4.00} & {\textbf{87.09}} \small{$\pm$0.85} & 97.86 \small{$\pm$0.17} & {72.14} \small{$\pm$2.23} & {\textbf{61.03}} \small{$\pm$8.55} & 49.22 \small{$\pm$5.12} & {78.01} \\
BitFit~\cite{Zaken2022bitfit} & 81.19 \small{$\pm$6.08} & 88.63 \small{$\pm$6.69} & {86.83} \small{$\pm$0.62} & 94.42 \small{$\pm$0.94} & 66.26 \small{$\pm$6.81} & 53.42 \small{$\pm$10.63} & 52.59 \small{$\pm$5.31} & 74.76 \\  
\midrule
\multicolumn{9}{c}{\textit{Gradient-Free Methods}} \\ 
\midrule
Manual Prompt & 79.82 & 89.65 & 76.96 & 41.33 & 67.40 & 31.11 & 51.62 & 62.56 \\
In-Context Learning~\cite{brown2020GPT3} & 79.79 \small{$\pm$3.06} & 85.38 \small{$\pm$3.92} & 62.21 \small{$\pm$13.46} & 34.83 \small{$\pm$7.59} & 45.81 \small{$\pm$6.67} & {47.11} \small{$\pm$0.63} & {60.36} \small{$\pm$1.56} & 59.36\\
\cdashline{1-9}
BBT~\cite{sun2022bbt} & {89.56} \small{$\pm$0.25} & {91.50} \small{$\pm$0.16} & {81.51} \small{$\pm$0.79} & 79.99 \small{$\pm$2.95} & 61.56 \small{$\pm$4.34} &  46.58 \small{$\pm$1.33}  & 52.59 \small{$\pm$2.21} & {71.90} \\
BBTv2~\cite{sun2022bbtv2} & 90.33 \small{$\pm$1.73} & 92.86 \small{$\pm$0.62} & {85.28} \small{$\pm$0.49} & 93.64 \small{$\pm$0.68} & 77.01 \small{$\pm$4.73} & 57.27 \small{$\pm$2.27} & {56.68} \small{$\pm$3.32} & 79.01 \\ 
\textbf{BBT-RGB (ours)} & \textbf{92.89} \small{$\pm$0.26} & \textbf{94.20} \small{$\pm$0.48} &  \textbf{85.60} \small{$\pm$0.41} &\textbf{94.41} \small{$\pm$0.73}  & \textbf{79.49} \small{$\pm$1.84} & \textbf{60.71} \small{$\pm$0.66}  & \textbf{61.82} \small{$\pm$1.20} & \textbf{81.30} \\
\bottomrule
\end{tabular}
}
\caption{Overall comparison between BBT-RGB and other methods~(both gradient-based and gradient-free). All of the results are obtained with the RoBERTa-Large~\cite{liu2019roberta} backbone in 16-shot (per class) setting.
The performance of the hitherto best technique combinations is reported in this table, 
as illustrated in Table~\ref{table:rgb-settings}. 
For each distinct combination, 
we demonstrate the ablation stduies in section~\ref{sec:ablations}.
}
\label{tab:main}
\vspace{-1em}
\end{table*}

\subsection{M\textsuperscript{2} Verbalizers}

Most prior works employ a single verbalizer for gradient-free optimization, which cannot make full use of the information, \textit{i.e.}, logits returned by the black box model.
To address this problem, 
we propose \textbf{M}ulti-\textbf{M}ixed verbalizers, 
which are constructed through the following methods: 
1) manual verbalizer selection\footnote{Specifically, we use synonyms in practice.}. 
2) search-based verbalizer construction based on word importance estimation by TF-IDF.
3) auto verbalizer generation based on neural nets~\cite{gao2021making}.
After verbalizers are selected by the aforementioned approaches, the confidence of each category is represented by the average prediction probability of multiple verbalizers.
Compared with the previous approach, 
M\textsuperscript{2} verbalizers make one step forward to exploit the information provided by the black-box model.
Additionally, this approach can prevent the negative impact on model performance caused by a single unsuitable label word.

\subsection{In\textsuperscript{2} Initialization}
An appropriate initialization has proven to play an essential role in effective prompt-based tuning~\cite{an2022input, prasad2022GrIPS}.
Inspired by previous efforts, we propose a model-agnostic strategy named as In\textsuperscript{2} initialization.
The first component of our approach is a task-specific manual \textbf{In}struction.
For the second part, we iterate through the training set and take each sample as a demonstration~\cite{sewon2022rethinking},
which is assessed on the validation set together with the pre-selected instruction. 
After that, the sample with the best performance is selected for \textbf{In}-context learning.

\section{Experiments}

\subsection{Experimental Settings}
\label{exp:settting}

\paragraph{Backbone}
We use RoBERTa-Large~\cite{liu2019roberta} as backbone throughout the experiments.


\paragraph{Datasets}
To evaluate our proposed methods, 
we choose a series of tasks from the GLUE benchmark~\cite{wang2018glue}.
Specifically, we employ SST-2~\cite{socher2013sst} and Yelp~\cite{zhang2015char} for sentiment analysis, AGNews and DBPedia~\cite{zhang2015char} for topic classification, SNLI~\cite{bowman2015snli} and RTE~\cite{dagan2005the} for natural language inference, and MRPC~\cite{dolan2005mrpc} for semantic paraphrasing.

\paragraph{Methods and Hyperparameters} For all the experiments we covered, 
the components of BBT-RGB we employed and hyperparameters are showcased in Table~\ref{table:rgb-settings} and Table~\ref{table:hparams}.
The details of experimental settings are listed in secion~\ref{sec:exp-details}.


\subsection{Main Results}

As is demonstrated in table~\ref{tab:main}, 
we compare BBT-RGB with both gradient-based and gradient-free tuning methods\footnote{We employ the results reported by~\citet{sun2022bbtv2} for comparison.}. 
We observed different levels of improvement on various NLP tasks.

\paragraph{Sentiment Analysis}
On both the SST-2 and Yelp datasets,
our method surpasses all prior white-box methods, 
consistently demonstrating superior performance compared to the established baselines.

\paragraph{Topic Classification}
Compared with the previous gradient-free method, BBT-RGB has a significant advancement in the evaluation based on DBPedia and AGNews but still needs to catch up to full model tuning.
We hold the view that this is caused by a relatively large number of classes (categories), and it is difficult for the model to learn enough knowledge under few-shot settings.

\paragraph{Entailment and Inference}
BBT-RGB benefits entailment and natural language inference tasks significantly; both experiments on SNLI and MRPC indicate surpassing full fine-tuning performance. 
In addition, we can observe a leap in the accuracy of RTE compared with previous baselines.

\subsection{Ablation Studies and Analysis}
\label{sec:ablations}

\paragraph{Ablation Studies.}
We conduct ablation studies to verify the effectiveness of the three proposed techniques that formed the core of this paper,
as demonstrated in Table~\ref{tab:ablations}. 
Overall, each component of BBT-RGB demonstrates gains across various tasks.

\begin{table}[ht]
\centering
\resizebox{\linewidth}{!}{
\begin{tabular}{lccc}
\toprule
\textbf{Method} & \textbf{SST-2} & \textbf{AG's News} & \textbf{RTE} \\ 
\midrule
\textbf{BBT-RGB} & 91.00     & 85.59        & 61.80      \\
\midrule
~w/o. M\textsuperscript{2}Verb    & 91.00      & 85.59      & 61.33      \\
~w/o. Two-Stage     & 90.39      & 85.57        & 61.17  \\ 
~w/o. In\textsuperscript{2}Init     & 90.77      & 84.31        & 60.17  \\
\midrule
~w/o. Two-Stage \& M\textsuperscript{2}Verb   & 90.83      & 85.56        & 59.77 \\ 
~w/o. Two-Stage \& In\textsuperscript{2}Init   & 90.28      & 83.79        & 59.30  \\ 
~w/o. In\textsuperscript{2}Init \& M\textsuperscript{2}Verb   & 90.66      & 83.75        & 59.47  \\ 

\bottomrule
\end{tabular}
}
\caption{Ablation studies of BBT-RGB on SST-2, AG's News, and RTE}
\label{tab:ablations}
\end{table}

To balance computational resource constraints and fairness,
we select one task each from Sentiment Analysis, Topic Classification, and Entailment \& Inference. 
Then we carry out ablation studies on each module comprising BBT-RGB (with \texttt{random seed = 42}).

\paragraph{Performance and Stability Comparison.}

\begin{figure}[ht]
    \centering
    \includegraphics[scale=0.38]{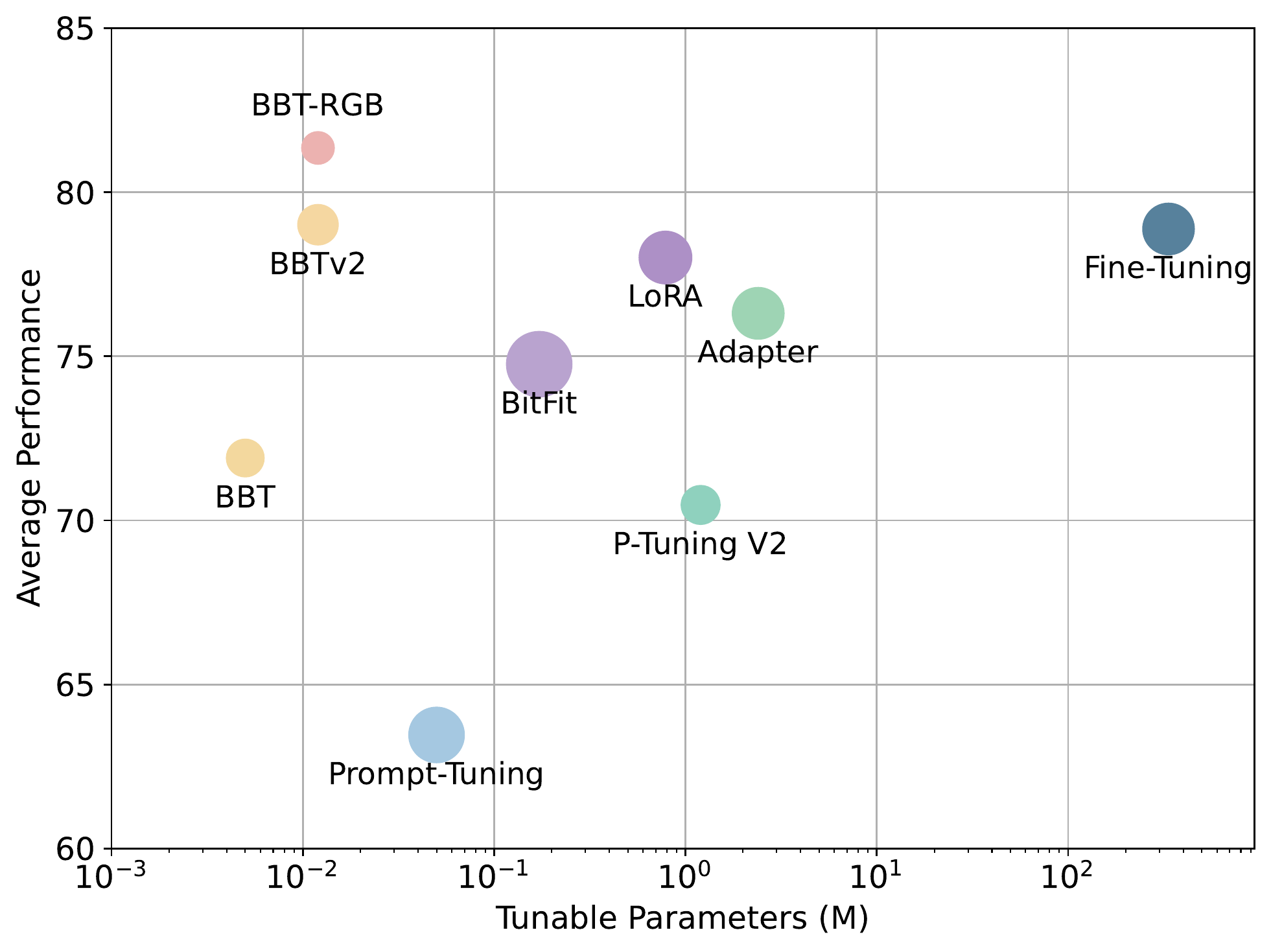}
    \caption{Comparing BBT-RGB with other tuning methods on average performance over seven tasks described in section~\ref{exp:settting}.
    The size of the circle is proportional to the standard deviation.}
    \label{fig:param-vis}
\end{figure}

Figure~\ref{fig:param-vis} illustrates a comparative analysis of BBT-RGB against other gradient-based and gradient-free methods in terms of performance, parameter tuning requirements, and stability. 

It is evident that while maintaining optimal performance, BBT-RGB incurs minimal computational overhead. Moreover, the standard deviation indicates BBT-RGB's superior stability compared to these methods, attributable to In\textsuperscript{2} Initialization enhancing the stability of few-shot learning.

Case studies pertaining to the optimization process can be found in section~\ref{appendix:case}. 

\section{Conclusion}
This paper proposes BBT-RGB, 
a set of simple but effective techniques to drive more powerful derivative-free prompt-based learning.
We make improvements from three independent aspects: 
(1) Two-stage derivative-free optimization algorithms for attenuating overfitting; 
(2) Versatile verbalizer construction with a robust selection process; 
(3) Using Instruction learning and demonstrations to exploit in-context information.
All the modules are ``plug-and-play'', 
and empirical studies across a series of tasks verify the effectiveness of our method.

\section*{Limitations and Ethical Consideration}

\paragraph{Limitations.}
Our limitations are threefold:
\begin{itemize}
    \item Following previous works~\citep{sun2022bbt, sun2022bbtv2},
our proposed method lays much emphasis on the optimization of continuous prompts.
It can be applied to a majority of open-source Large Language Models (LLMs), but for some commercial models that do not provide loss, logits, or perplexity, the optimization is constrained to remain in the discrete form at the initial layer of the model.
    \item Since the algorithm is unable to achieve linear convergence,
some of the tasks require more API calls,
which may lead to extra costs when running on commercial models.
    \item Given that In\textsuperscript{2}Init and M\textsuperscript{2}Verb involve the search for verbalizers and demonstrations, our method takes a longer execution time compared to BBTv2, requiring approximately 25\% additional runtime.
\end{itemize}

\paragraph*{Ethical Considerations.}
Our method: BBT-RGB, aims to exploit the potential of black-box tuning further, and the contribution in this paper is fully methodological. 
Therefore, this contribution has no direct negative social or ethical impacts.
Moreover, 
given that our approach requires significantly less computational resources compared to full-fine tuning, 
it is poised to contribute positively to the sustainable development of the community.

\section*{Acknowledgment}
This work is supported by Shanghai ``Science and Technology Innovation Action Plan'' Project (No.23511100700).
Our method is also derived from a prize-winning solution of the \textit{First International Algorithm Case Competition: PLM Tuning Track, Guangdong-Hong Kong-Macao Greater Bay Area}.
Finally, 
we thank our anonymous reviewers for their insightful comments and suggestions.

\bibliography{anthology,custom}

\begin{thebibliography}{46}
\expandafter\ifx\csname natexlab\endcsname\relax\def\natexlab#1{#1}\fi

\bibitem[{Aghajanyan et~al.(2021)Aghajanyan, Gupta, and
  Zettlemoyer}]{aghajanyan2021intrinsic}
Armen Aghajanyan, Sonal Gupta, and Luke Zettlemoyer. 2021.
\newblock \href {https://doi.org/10.18653/v1/2021.acl-long.568} {Intrinsic
  dimensionality explains the effectiveness of language model fine-tuning}.
\newblock In \emph{Proceedings of the 59th Annual Meeting of the Association
  for Computational Linguistics and the 11th International Joint Conference on
  Natural Language Processing, {ACL/IJCNLP} 2021, (Volume 1: Long Papers),
  Virtual Event, August 1-6, 2021}, pages 7319--7328. Association for
  Computational Linguistics.

\bibitem[{An et~al.(2022)An, Li, Lin, Liu, Chen, Fu, Chen, Zheng, and
  Lou}]{an2022input}
Shengnan An, Yifei Li, Zeqi Lin, Qian Liu, Bei Chen, Qiang Fu, Weizhu Chen,
  Nanning Zheng, and Jian-Guang Lou. 2022.
\newblock \href {https://doi.org/10.48550/ARXIV.2203.03131} {Input-tuning:
  Adapting unfamiliar inputs to frozen pretrained models}.

\bibitem[{Ben~Zaken et~al.(2022)Ben~Zaken, Goldberg, and
  Ravfogel}]{Zaken2022bitfit}
Elad Ben~Zaken, Yoav Goldberg, and Shauli Ravfogel. 2022.
\newblock \href {https://doi.org/10.18653/v1/2022.acl-short.1} {{B}it{F}it:
  Simple parameter-efficient fine-tuning for transformer-based masked
  language-models}.
\newblock In \emph{Proceedings of the 60th Annual Meeting of the Association
  for Computational Linguistics (Volume 2: Short Papers)}, pages 1--9, Dublin,
  Ireland. Association for Computational Linguistics.

\bibitem[{Bowman et~al.(2015)Bowman, Angeli, Potts, and
  Manning}]{bowman2015snli}
Samuel~R. Bowman, Gabor Angeli, Christopher Potts, and Christopher~D. Manning.
  2015.
\newblock \href {https://doi.org/10.18653/v1/d15-1075} {A large annotated
  corpus for learning natural language inference}.
\newblock In \emph{Proceedings of the 2015 Conference on Empirical Methods in
  Natural Language Processing, {EMNLP} 2015, Lisbon, Portugal, September 17-21,
  2015}, pages 632--642. The Association for Computational Linguistics.

\bibitem[{Brown et~al.(2020)Brown, Mann, Ryder, Subbiah, Kaplan, Dhariwal,
  Neelakantan, Shyam, Sastry, Askell, Agarwal, Herbert-Voss, Krueger, Henighan,
  Child, Ramesh, Ziegler, Wu, Winter, Hesse, Chen, Sigler, Litwin, Gray, Chess,
  Clark, Berner, McCandlish, Radford, Sutskever, and Amodei}]{brown2020GPT3}
Tom Brown, Benjamin Mann, Nick Ryder, Melanie Subbiah, Jared~D Kaplan, Prafulla
  Dhariwal, Arvind Neelakantan, Pranav Shyam, Girish Sastry, Amanda Askell,
  Sandhini Agarwal, Ariel Herbert-Voss, Gretchen Krueger, Tom Henighan, Rewon
  Child, Aditya Ramesh, Daniel Ziegler, Jeffrey Wu, Clemens Winter, Chris
  Hesse, Mark Chen, Eric Sigler, Mateusz Litwin, Scott Gray, Benjamin Chess,
  Jack Clark, Christopher Berner, Sam McCandlish, Alec Radford, Ilya Sutskever,
  and Dario Amodei. 2020.
\newblock \href
  {https://proceedings.neurips.cc/paper/2020/file/1457c0d6bfcb4967418bfb8ac142f64a-Paper.pdf}
  {Language models are few-shot learners}.
\newblock In \emph{Advances in Neural Information Processing Systems},
  volume~33, pages 1877--1901. Curran Associates, Inc.

\bibitem[{Chai et~al.(2022)Chai, Wang, Sun, Tian, Wu, and Wang}]{chai2022clip}
Yekun Chai, Shuohuan Wang, Yu~Sun, Hao Tian, Hua Wu, and Haifeng Wang. 2022.
\newblock \href {https://aclanthology.org/2022.findings-emnlp.8} {Clip-tuning:
  Towards derivative-free prompt learning with a mixture of rewards}.
\newblock In \emph{Findings of the Association for Computational Linguistics:
  EMNLP 2022}, pages 108--117, Abu Dhabi, United Arab Emirates. Association for
  Computational Linguistics.

\bibitem[{Chen et~al.(2021)Chen, Tworek, Jun, Yuan, Ponde, Kaplan, Edwards,
  Burda, Joseph, Brockman et~al.}]{chen2021evaluating}
Mark Chen, Jerry Tworek, Heewoo Jun, Qiming Yuan, Henrique Ponde, Jared Kaplan,
  Harri Edwards, Yura Burda, Nicholas Joseph, Greg Brockman, et~al. 2021.
\newblock Evaluating large language models trained on code.
\newblock \emph{arXiv preprint arXiv:2107.03374}.

\bibitem[{Chowdhery et~al.(2022)Chowdhery, Narang, Devlin, Bosma, Mishra,
  Roberts, Barham, Chung, Sutton, Gehrmann et~al.}]{chowdhery2022palm}
Aakanksha Chowdhery, Sharan Narang, Jacob Devlin, Maarten Bosma, Gaurav Mishra,
  Adam Roberts, Paul Barham, Hyung~Won Chung, Charles Sutton, Sebastian
  Gehrmann, et~al. 2022.
\newblock Palm: Scaling language modeling with pathways.

\bibitem[{Dagan et~al.(2005)Dagan, Glickman, and Magnini}]{dagan2005the}
Ido Dagan, Oren Glickman, and Bernardo Magnini. 2005.
\newblock The {PASCAL} recognising textual entailment challenge.
\newblock In \emph{{MLCW}}, volume 3944 of \emph{Lecture Notes in Computer
  Science}, pages 177--190. Springer.

\bibitem[{Devlin et~al.(2019)Devlin, Chang, Lee, and
  Toutanova}]{devlin2018bert}
Jacob Devlin, Ming-Wei Chang, Kenton Lee, and Kristina Toutanova. 2019.
\newblock \href {https://doi.org/10.18653/v1/N19-1423} {{BERT}: Pre-training of
  deep bidirectional transformers for language understanding}.
\newblock In \emph{Proceedings of the 2019 Conference of the North {A}merican
  Chapter of the Association for Computational Linguistics: Human Language
  Technologies, Volume 1 (Long and Short Papers)}, pages 4171--4186,
  Minneapolis, Minnesota. Association for Computational Linguistics.

\bibitem[{Diao et~al.(2023)Diao, Huang, Xu, Li, Yong, Zhou, and
  Zhang}]{Diao2022bbp}
Shizhe Diao, Zhichao Huang, Ruijia Xu, Xuechun Li, LIN Yong, Xiao Zhou, and
  Tong Zhang. 2023.
\newblock \href {https://openreview.net/forum?id=IvsGP7xRvm} {Black-box prompt
  learning for pre-trained language models}.
\newblock \emph{Transactions on Machine Learning Research}.

\bibitem[{Dolan and Brockett(2005)}]{dolan2005mrpc}
William~B. Dolan and Chris Brockett. 2005.
\newblock \href {https://aclanthology.org/I05-5002/} {Automatically
  constructing a corpus of sentential paraphrases}.
\newblock In \emph{Proceedings of the Third International Workshop on
  Paraphrasing, IWP@IJCNLP 2005, Jeju Island, Korea, October 2005, 2005}. Asian
  Federation of Natural Language Processing.

\bibitem[{Gao et~al.(2021)Gao, Fisch, and Chen}]{gao2021making}
Tianyu Gao, Adam Fisch, and Danqi Chen. 2021.
\newblock \href {https://doi.org/10.18653/v1/2021.acl-long.295} {Making
  pre-trained language models better few-shot learners}.
\newblock In \emph{Proceedings of the 59th Annual Meeting of the Association
  for Computational Linguistics and the 11th International Joint Conference on
  Natural Language Processing (Volume 1: Long Papers)}, pages 3816--3830,
  Online. Association for Computational Linguistics.

\bibitem[{Han et~al.(2023)Han, Cui, Zhu, Wang, Chen, Sun, Li, and
  Gao}]{han2023gdfo}
Chengcheng Han, Liqing Cui, Renyu Zhu, Jianing Wang, Nuo Chen, Qiushi Sun,
  Xiang Li, and Ming Gao. 2023.
\newblock \href {https://doi.org/10.18653/v1/2023.findings-acl.55} {When
  gradient descent meets derivative-free optimization: A match made in
  black-box scenario}.
\newblock In \emph{Findings of the Association for Computational Linguistics:
  ACL 2023}, pages 868--880, Toronto, Canada. Association for Computational
  Linguistics.

\bibitem[{Hansen et~al.(2003)Hansen, M{\"{u}}ller, and
  Koumoutsakos}]{hansen2003reducing}
Nikolaus Hansen, Sibylle~D. M{\"{u}}ller, and Petros Koumoutsakos. 2003.
\newblock \href {https://doi.org/10.1162/106365603321828970} {Reducing the time
  complexity of the derandomized evolution strategy with covariance matrix
  adaptation {(CMA-ES)}}.
\newblock \emph{Evol. Comput.}, 11(1):1--18.

\bibitem[{Hansen and Ostermeier(2001)}]{hansen2001cma}
Nikolaus Hansen and Andreas Ostermeier. 2001.
\newblock \href {https://doi.org/10.1162/106365601750190398} {Completely
  derandomized self-adaptation in evolution strategies}.
\newblock \emph{Evol. Comput.}, 9(2):159--195.

\bibitem[{Hou et~al.(2023)Hou, O'Connor, Andreas, Chang, and
  Zhang}]{hou2022promptboosting}
Bairu Hou, Joe O'Connor, Jacob Andreas, Shiyu Chang, and Yang Zhang. 2023.
\newblock \href {https://proceedings.mlr.press/v202/hou23b.html}
  {{P}rompt{B}oosting: Black-box text classification with ten forward passes}.
\newblock In \emph{Proceedings of the 40th International Conference on Machine
  Learning}, volume 202 of \emph{Proceedings of Machine Learning Research},
  pages 13309--13324. PMLR.

\bibitem[{Houlsby et~al.(2019)Houlsby, Giurgiu, Jastrzebski, Morrone,
  De~Laroussilhe, Gesmundo, Attariyan, and Gelly}]{houlsby2019ParamEff}
Neil Houlsby, Andrei Giurgiu, Stanislaw Jastrzebski, Bruna Morrone, Quentin
  De~Laroussilhe, Andrea Gesmundo, Mona Attariyan, and Sylvain Gelly. 2019.
\newblock \href {https://proceedings.mlr.press/v97/houlsby19a.html}
  {Parameter-efficient transfer learning for {NLP}}.
\newblock In \emph{Proceedings of the 36th International Conference on Machine
  Learning}, volume~97 of \emph{Proceedings of Machine Learning Research},
  pages 2790--2799. PMLR.

\bibitem[{Hu et~al.(2022)Hu, yelong shen, Wallis, Allen-Zhu, Li, Wang, Wang,
  and Chen}]{hu2022lora}
Edward~J Hu, yelong shen, Phillip Wallis, Zeyuan Allen-Zhu, Yuanzhi Li, Shean
  Wang, Lu~Wang, and Weizhu Chen. 2022.
\newblock \href {https://openreview.net/forum?id=nZeVKeeFYf9} {Lo{RA}: Low-rank
  adaptation of large language models}.
\newblock In \emph{International Conference on Learning Representations}.

\bibitem[{Lester et~al.(2021)Lester, Al-Rfou, and Constant}]{lester2021power}
Brian Lester, Rami Al-Rfou, and Noah Constant. 2021.
\newblock \href {https://doi.org/10.18653/v1/2021.emnlp-main.243} {The power of
  scale for parameter-efficient prompt tuning}.
\newblock In \emph{Proceedings of the 2021 Conference on Empirical Methods in
  Natural Language Processing}, pages 3045--3059, Online and Punta Cana,
  Dominican Republic. Association for Computational Linguistics.

\bibitem[{Li and Liang(2021)}]{li2021prefix}
Xiang~Lisa Li and Percy Liang. 2021.
\newblock \href {https://doi.org/10.18653/v1/2021.acl-long.353} {Prefix-tuning:
  Optimizing continuous prompts for generation}.
\newblock In \emph{Proceedings of the 59th Annual Meeting of the Association
  for Computational Linguistics and the 11th International Joint Conference on
  Natural Language Processing (Volume 1: Long Papers)}, pages 4582--4597,
  Online. Association for Computational Linguistics.

\bibitem[{Lin et~al.(2022)Lin, Wang, Liu, and Qiu}]{lin2022transformer}
Tianyang Lin, Yuxin Wang, Xiangyang Liu, and Xipeng Qiu. 2022.
\newblock \href {https://doi.org/https://doi.org/10.1016/j.aiopen.2022.10.001}
  {A survey of transformers}.
\newblock \emph{AI Open}, 3:111--132.

\bibitem[{Liu et~al.(2023)Liu, Yuan, Fu, Jiang, Hayashi, and
  Neubig}]{liu2021promptsurvey}
Pengfei Liu, Weizhe Yuan, Jinlan Fu, Zhengbao Jiang, Hiroaki Hayashi, and
  Graham Neubig. 2023.
\newblock \href {https://doi.org/10.1145/3560815} {Pre-train, prompt, and
  predict: A systematic survey of prompting methods in natural language
  processing}.
\newblock \emph{ACM Comput. Surv.}, 55(9).

\bibitem[{Liu et~al.(2022)Liu, Ji, Fu, Tam, Du, Yang, and
  Tang}]{liu2022ptuning}
Xiao Liu, Kaixuan Ji, Yicheng Fu, Weng Tam, Zhengxiao Du, Zhilin Yang, and Jie
  Tang. 2022.
\newblock \href {https://doi.org/10.18653/v1/2022.acl-short.8} {{P}-tuning:
  Prompt tuning can be comparable to fine-tuning across scales and tasks}.
\newblock In \emph{Proceedings of the 60th Annual Meeting of the Association
  for Computational Linguistics (Volume 2: Short Papers)}, pages 61--68,
  Dublin, Ireland. Association for Computational Linguistics.

\bibitem[{Liu et~al.(2019)Liu, Ott, Goyal, Du, Joshi, Chen, Levy, Lewis,
  Zettlemoyer, and Stoyanov}]{liu2019roberta}
Yinhan Liu, Myle Ott, Naman Goyal, Jingfei Du, Mandar Joshi, Danqi Chen, Omer
  Levy, Mike Lewis, Luke Zettlemoyer, and Veselin Stoyanov. 2019.
\newblock {RoBERTa}: A robustly optimized {BERT} pretraining approach.
\newblock \emph{arXiv preprint arXiv:1907.11692}.

\bibitem[{Min et~al.(2022)Min, Lyu, Holtzman, Artetxe, Lewis, Hajishirzi, and
  Zettlemoyer}]{sewon2022rethinking}
Sewon Min, Xinxi Lyu, Ari Holtzman, Mikel Artetxe, Mike Lewis, Hannaneh
  Hajishirzi, and Luke Zettlemoyer. 2022.
\newblock \href {https://doi.org/10.48550/ARXIV.2202.12837} {Rethinking the
  role of demonstrations: What makes in-context learning work?}

\bibitem[{OpenAI(2023)}]{openai2023gpt4}
OpenAI. 2023.
\newblock \href {http://arxiv.org/abs/2303.08774} {{GPT-4} technical report}.

\bibitem[{Ouyang et~al.(2022)Ouyang, Wu, Jiang, Almeida, Wainwright, Mishkin,
  Zhang, Agarwal, Slama, Ray, Schulman, Hilton, Kelton, Miller, Simens, Askell,
  Welinder, Christiano, Leike, and Lowe}]{ouyang2022trainingLM}
Long Ouyang, Jeff Wu, Xu~Jiang, Diogo Almeida, Carroll~L. Wainwright, Pamela
  Mishkin, Chong Zhang, Sandhini Agarwal, Katarina Slama, Alex Ray, John
  Schulman, Jacob Hilton, Fraser Kelton, Luke~E. Miller, Maddie Simens, Amanda
  Askell, Peter Welinder, Paul~Francis Christiano, Jan Leike, and Ryan~J. Lowe.
  2022.
\newblock \href {https://arxiv.org/abs/2203.02155} {Training language models to
  follow instructions with human feedback}.
\newblock {arXiv:2203.02155}.

\bibitem[{Powell(1994)}]{powell1994direct}
M.~J.~D. Powell. 1994.
\newblock \href {https://doi.org/10.1007/978-94-015-8330-5_4} {A direct search
  optimization method that models the objective and constraint functions by
  linear interpolation}.
\newblock \emph{Advances in Optimization and Numerical Analysis}, pages 51--67.

\bibitem[{Powell(1998)}]{powell1998direct}
M.~J.~D. Powell. 1998.
\newblock \href {https://doi.org/10.1017/S0962492900002841} {Direct search
  algorithms for optimization calculations}.
\newblock \emph{Acta Numerica}, 7:287–336.

\bibitem[{Prasad et~al.(2022)Prasad, Hase, Zhou, and Bansal}]{prasad2022GrIPS}
Archiki Prasad, Peter Hase, Xiang Zhou, and Mohit Bansal. 2022.
\newblock \href {https://doi.org/10.48550/ARXIV.2203.07281} {Grips:
  Gradient-free, edit-based instruction search for prompting large language
  models}.

\bibitem[{Qian et~al.(2016)Qian, Hu, and Yu}]{qian2016dfo}
Hong Qian, Yi{-}Qi Hu, and Yang Yu. 2016.
\newblock \href {http://www.ijcai.org/Abstract/16/278} {Derivative-free
  optimization of high-dimensional non-convex functions by sequential random
  embeddings}.
\newblock In \emph{Proceedings of the Twenty-Fifth International Joint
  Conference on Artificial Intelligence, {IJCAI} 2016, New York, NY, USA, 9-15
  July 2016}, pages 1946--1952. {IJCAI/AAAI} Press.

\bibitem[{Qin and Eisner(2021)}]{qin2021learning}
Guanghui Qin and Jason Eisner. 2021.
\newblock \href {https://doi.org/10.18653/v1/2021.naacl-main.410} {Learning how
  to ask: Querying {LM}s with mixtures of soft prompts}.
\newblock In \emph{Proceedings of the 2021 Conference of the North American
  Chapter of the Association for Computational Linguistics: Human Language
  Technologies}, pages 5203--5212, Online. Association for Computational
  Linguistics.

\bibitem[{Qiu et~al.(2020)Qiu, Sun, Xu, Shao, Dai, and Huang}]{qiu2020pre}
Xipeng Qiu, Tianxiang Sun, Yige Xu, Yunfan Shao, Ning Dai, and Xuanjing Huang.
  2020.
\newblock \href {https://doi.org/10.1007/s11431-020-1647-3} {Pre-trained models
  for natural language processing: A survey}.
\newblock \emph{Science China Technological Sciences}, 63(10):1872--1897.

\bibitem[{Rios and Sahinidis(2013)}]{rios2013DFOreview}
Luis~Miguel Rios and Nikolaos~V. Sahinidis. 2013.
\newblock Derivative-free optimization: {A} review of algorithms and comparison
  of software implementations.
\newblock \emph{Journal of Global Optimization}, 56(3):1247--1293.

\bibitem[{Scao et~al.(2022)Scao, Fan, Akiki, Pavlick, Ili{\'c}, Hesslow,
  Castagn{\'e}, Luccioni, Yvon, Gall{\'e} et~al.}]{scao2022bloom}
Teven~Le Scao, Angela Fan, Christopher Akiki, Ellie Pavlick, Suzana Ili{\'c},
  Daniel Hesslow, Roman Castagn{\'e}, Alexandra~Sasha Luccioni, Fran{\c{c}}ois
  Yvon, Matthias Gall{\'e}, et~al. 2022.
\newblock \href {https://arxiv.org/abs/2211.05100} {Bloom: A 176b-parameter
  open-access multilingual language model}.
\newblock \emph{ArXiv preprint}, abs/2211.05100.

\bibitem[{Schick et~al.(2020)Schick, Schmid, and
  Sch{\"u}tze}]{schick2020automatically}
Timo Schick, Helmut Schmid, and Hinrich Sch{\"u}tze. 2020.
\newblock \href {https://doi.org/10.18653/v1/2020.coling-main.488}
  {Automatically identifying words that can serve as labels for few-shot text
  classification}.
\newblock In \emph{Proceedings of the 28th International Conference on
  Computational Linguistics}, pages 5569--5578, Barcelona, Spain (Online).
  International Committee on Computational Linguistics.

\bibitem[{Socher et~al.(2013)Socher, Perelygin, Wu, Chuang, Manning, Ng, and
  Potts}]{socher2013sst}
Richard Socher, Alex Perelygin, Jean Wu, Jason Chuang, Christopher~D. Manning,
  Andrew~Y. Ng, and Christopher Potts. 2013.
\newblock \href {https://aclanthology.org/D13-1170/} {Recursive deep models for
  semantic compositionality over a sentiment treebank}.
\newblock In \emph{Proceedings of the 2013 Conference on Empirical Methods in
  Natural Language Processing, {EMNLP} 2013, 18-21 October 2013, Grand Hyatt
  Seattle, Seattle, Washington, USA, {A} meeting of SIGDAT, a Special Interest
  Group of the {ACL}}, pages 1631--1642. {ACL}.

\bibitem[{Sun et~al.(2022{\natexlab{a}})Sun, He, Qian, Zhou, Huang, and
  Qiu}]{sun2022bbtv2}
Tianxiang Sun, Zhengfu He, Hong Qian, Yunhua Zhou, Xuanjing Huang, and Xipeng
  Qiu. 2022{\natexlab{a}}.
\newblock \href {https://aclanthology.org/2022.emnlp-main.259} {{BBT}v2:
  Towards a gradient-free future with large language models}.
\newblock In \emph{Proceedings of the 2022 Conference on Empirical Methods in
  Natural Language Processing}, pages 3916--3930, Abu Dhabi, United Arab
  Emirates. Association for Computational Linguistics.

\bibitem[{Sun et~al.(2022{\natexlab{b}})Sun, Shao, Qian, Huang, and
  Qiu}]{sun2022bbt}
Tianxiang Sun, Yunfan Shao, Hong Qian, Xuanjing Huang, and Xipeng Qiu.
  2022{\natexlab{b}}.
\newblock \href {https://arxiv.org/abs/2201.03514} {Black-box tuning for
  language-model-as-a-service}.
\newblock In \emph{Proceedings of the 39th International Conference on Machine
  Learning, {ICML} 2022, Baltimore, Maryland, USA}.

\bibitem[{Touvron et~al.(2023)Touvron, Lavril, Izacard, Martinet, Lachaux,
  Lacroix, Rozi{\`e}re, Goyal, Hambro, Azhar, Rodriguez, Joulin, Grave, and
  Lample}]{touvron2023llama}
Hugo Touvron, Thibaut Lavril, Gautier Izacard, Xavier Martinet, Marie-Anne
  Lachaux, Timoth{\'e}e Lacroix, Baptiste Rozi{\`e}re, Naman Goyal, Eric
  Hambro, Faisal Azhar, Aurelien Rodriguez, Armand Joulin, Edouard Grave, and
  Guillaume Lample. 2023.
\newblock Llama: Open and efficient foundation language models.
\newblock \emph{arXiv preprint arXiv:2302.13971}.

\bibitem[{Vaswani et~al.(2017)Vaswani, Shazeer, Parmar, Uszkoreit, Jones,
  Gomez, Kaiser, and Polosukhin}]{vaswani2017attention}
Ashish Vaswani, Noam Shazeer, Niki Parmar, Jakob Uszkoreit, Llion Jones,
  Aidan~N Gomez, \L~ukasz Kaiser, and Illia Polosukhin. 2017.
\newblock \href
  {https://proceedings.neurips.cc/paper/2017/file/3f5ee243547dee91fbd053c1c4a845aa-Paper.pdf}
  {Attention is all you need}.
\newblock In \emph{Advances in Neural Information Processing Systems},
  volume~30. Curran Associates, Inc.

\bibitem[{Wang et~al.(2018)Wang, Singh, Michael, Hill, Levy, and
  Bowman}]{wang2018glue}
Alex Wang, Amanpreet Singh, Julian Michael, Felix Hill, Omer Levy, and Samuel
  Bowman. 2018.
\newblock \href {https://doi.org/10.18653/v1/W18-5446} {{GLUE}: A multi-task
  benchmark and analysis platform for natural language understanding}.
\newblock In \emph{Proceedings of the 2018 {EMNLP} Workshop {B}lackbox{NLP}:
  Analyzing and Interpreting Neural Networks for {NLP}}, pages 353--355,
  Brussels, Belgium. Association for Computational Linguistics.

\bibitem[{Wierstra et~al.(2014)Wierstra, Schaul, Glasmachers, Sun, Peters, and
  Schmidhuber}]{wierstra2014nes}
Daan Wierstra, Tom Schaul, Tobias Glasmachers, Yi~Sun, Jan Peters, and
  J\"{u}rgen Schmidhuber. 2014.
\newblock \href {https://dl.acm.org/doi/10.5555/2627435.2638566} {Natural
  evolution strategies}.
\newblock \emph{Journal of Machine Learning Research}, 15(1):949–980.

\bibitem[{Zhang et~al.(2022)Zhang, Roller, Goyal, Artetxe, Chen, Chen, Dewan,
  Diab, Li, Lin, Mihaylov, Ott, Shleifer, Shuster, Simig, Koura, Sridhar, Wang,
  and Zettlemoyer}]{zhang2022opt}
Susan Zhang, Stephen Roller, Naman Goyal, Mikel Artetxe, Moya Chen, Shuohui
  Chen, Christopher Dewan, Mona Diab, Xian Li, Xi~Victoria Lin, Todor Mihaylov,
  Myle Ott, Sam Shleifer, Kurt Shuster, Daniel Simig, Punit~Singh Koura, Anjali
  Sridhar, Tianlu Wang, and Luke Zettlemoyer. 2022.
\newblock \href {http://arxiv.org/abs/2205.01068} {Opt: Open pre-trained
  transformer language models}.

\bibitem[{Zhang et~al.(2015)Zhang, Zhao, and LeCun}]{zhang2015char}
Xiang Zhang, Junbo~Jake Zhao, and Yann LeCun. 2015.
\newblock \href
  {https://proceedings.neurips.cc/paper/2015/hash/250cf8b51c773f3f8dc8b4be867a9a02-Abstract.html}
  {Character-level convolutional networks for text classification}.
\newblock In \emph{Advances in Neural Information Processing Systems 28: Annual
  Conference on Neural Information Processing Systems 2015, December 7-12,
  2015, Montreal, Quebec, Canada}, pages 649--657.

\end{thebibliography}
\bibliographystyle{acl_natbib}

\appendix


\section{Derivative-Free Prompt Tuning}
\label{sec:bbt}
Given a batch of samples $(X, Y)$ converted with prompt templates and label words,
the original derivative-free prompt learning, as introduced by~\citet{sun2022bbtv2} first use a set of prompt embeddings $p$ to concatenate the input tokens, 
creating the prompted input for LLMs with frozen backbones.
The prompt $p = p_0 + p_\theta$ consists of the initial prompt $p_0 \in \mathbb{R}^D$, which is manually/randomly selected and a tunable prompt $p_\theta \in \mathbb{R}^{D}$ that is progressively optimized through a DFO algorithm like CMA-ES~\citep{hansen2003reducing}. 
DFOs suffer slow convergence on high-dimensional problems,
but fortunately, \citet{aghajanyan2021intrinsic} discover that PLMs exhibit low-dimensional reparameterization that is as effective for fine-tuning as the full parameter space. 
This finding indicates that the search space of $p_\theta$ can be condensed into an intrinsic dimensionality ${z} \in \mathbb{R}^d ~(d \ll D)$ by using a (frozen) random projection matrix $\Pi\in\mathbb{R}^{D \times d}$, 
such that $p_\theta = \Pi \cdot {z} $ will
significantly decrease the cost of optimization.
Subsequently, the task-specific inference of model $f$ through API Call is performed to determine the fitness of candidate prompts using an objective function $\mathcal{L}(f([P;X]), Y)$, where $\mathcal{L}$ is a loss function such as cross-entropy.
Finally, 
the DFO algorithm iteratively refines the prompt for seeking ${p^{*}} = \arg\min_{p} \mathcal{L} (f([P;X]), Y)$.


In the era of large language models, 
black-box optimization is a promising research target that can drive models for few-shot learning without access to gradients. 
\citet{sun2022bbt} first propose BBT that focuses on optimizing continuous prompt by only accessing inference APIs 
and then present BBTv2~\cite{sun2022bbtv2} as an improved version.
While some recent works
focus on optimizing discrete prompts concurrent with our work.
\citet{Diao2022bbp} present black-box discrete prompt learning with gradient estimation as their key feature.
\citet{hou2022promptboosting} first use gradient-free methods to sample sub-optimal discrete prompts and then ensemble them by boosting algorithm. And \citet{chai2022clip} acquire informative feedback to enhance derivative-free optimization through using frozen subnetworks as critics.
Recently, \citet{han2023gdfo} ingeniously leverage knowledge distillation to combine gradient descent and gradient-free optimization.




\section{Detailed Two-Stage DFOs Algorithm}
\label{sec:two-stage-detailed}

Algorithm~\ref{alg:two-stage-detailed} is the full version of the Two-Stage DFOs we used in this paper. For the Evolutionary algorithm and the search-based algorithm, we select CMA-ES and COBYLA, respectively.

\begin{algorithm}[htb]
	\caption{Two-Stage DFOs (Detailed)}
	\label{alg:two-stage-detailed}
	\begin{algorithmic}[1]
		\Require \
		 popsize: $\lambda$, intrinsic dimension: $d$
            \Require \
          budget1:$b1$, budget2:$b2$, backbone:$f_{model}$
            \Require \
		 $m$, $\delta$, $C$, $D$  
     // initialize state variables
		\Ensure \
		hidden variable: $z$
		\Function{Two-Stage DFO}{}   
            \Statex\ // CMA-ES
            \Repeat
            \For{each hidden layer}
            \For{$i$ in 1 to $\lambda$}
            \State Sample $z_i$ from $\mathcal{N}$($m$,$\delta^2C$) 
            \State $f_i=f_{model}(z_i)$
            \EndFor
            \State Update $m, \delta, C$ with $f$
            \EndFor
            \State Update $z$ to min(f)
            \Until{ $b1$ times $f_{model}$ call}
            \Statex\ // COBYLA
            \For{each hidden layer}
            \Repeat
            \For{each search direction i in D}
            \State Update $z$ to min(f) along i
            \EndFor
            \State Select a new set of D
            \Until{ $b2//d$ times $f_{model}$ call}
            
            \EndFor
	    \EndFunction
	\end{algorithmic}
\end{algorithm}

\section{Case Studies}
\label{appendix:case}

\begin{figure}[ht]
    \centering
    \subfigure[CMA-ES]{
	\includegraphics[width=0.231\textwidth]{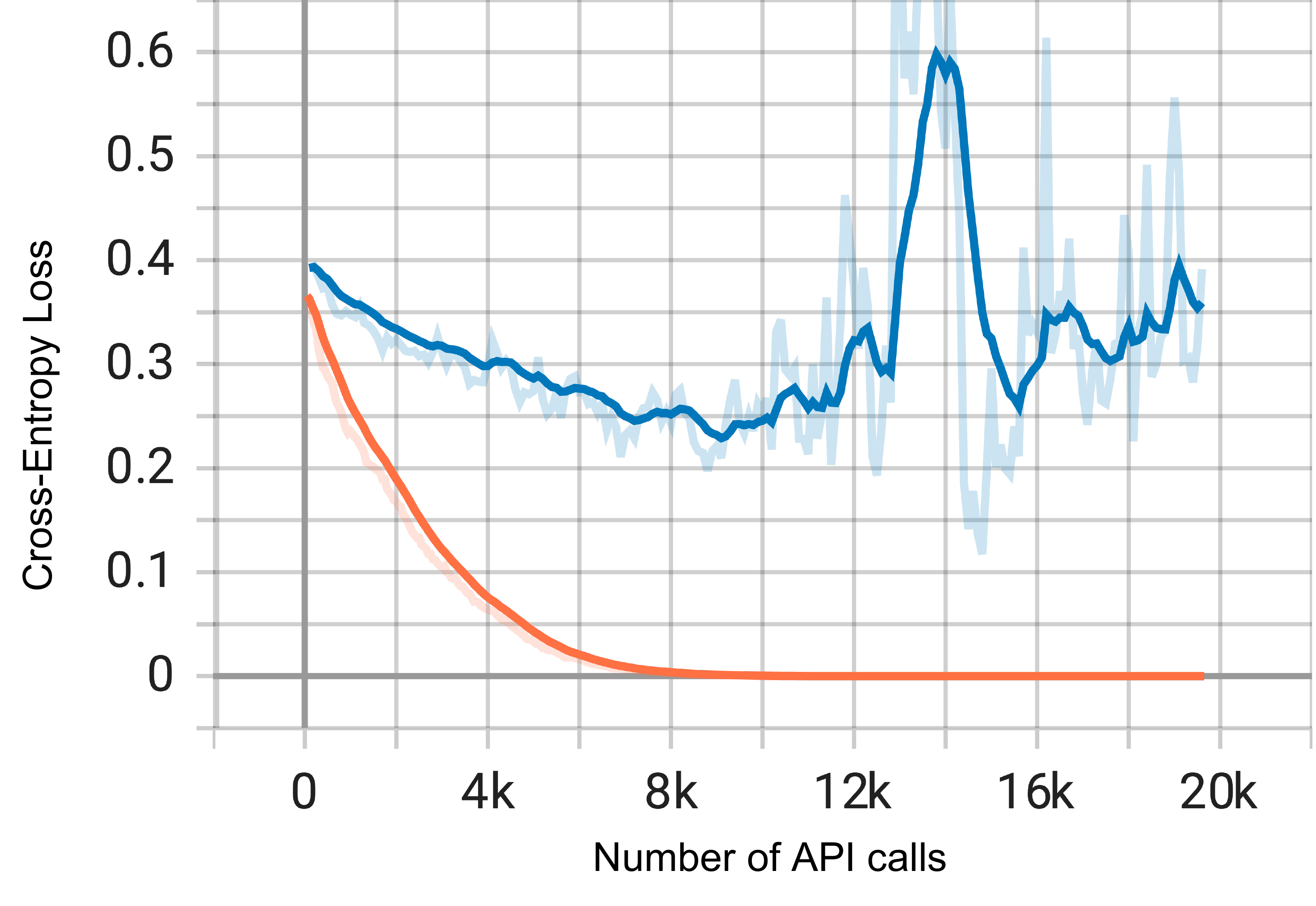}
	\label{pic1:yelp-no2stage}
    }
    \subfigure[Two-Stage DFOs]{
	\includegraphics[width=0.22\textwidth]{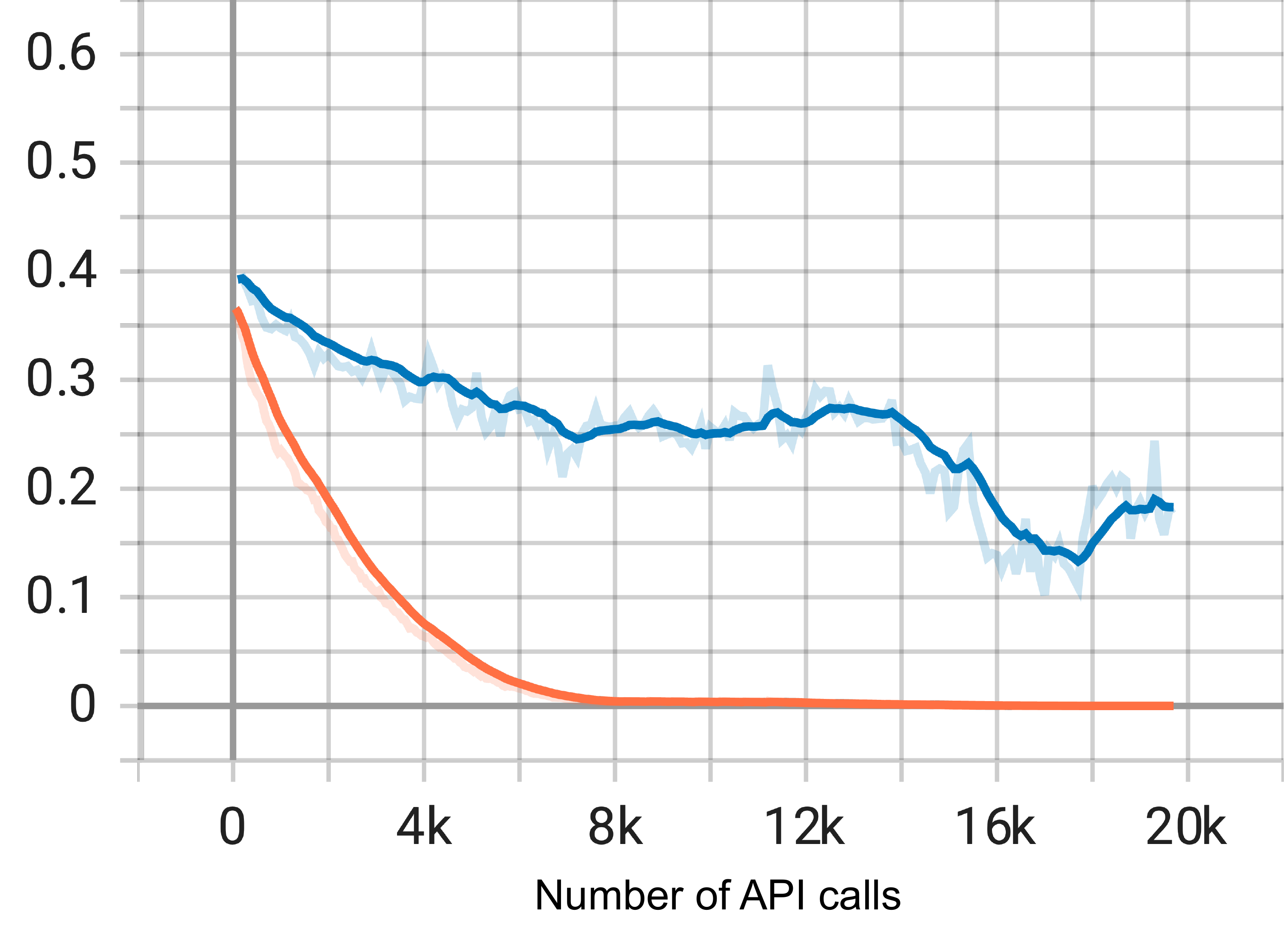}
	\label{pic1:yelp-2stage}
    }
    \caption{Comparison of original CMA-ES and Two-stage DFOs on Yelp dataset.}
    \label{fig:2stage-comp}
\end{figure}

We select two cases\footnote{We choose CMA-ES (8,000 budgets) and COBYLA (12,000 budgets) for the two-stage DFOs in this illustration.} 
to analyze the effectiveness of two-stage DFOs on Yelp dataset. 
In Figure~\ref{fig:2stage-comp}, 
the training loss (orange curve) converges to zero for both methods.
While the oscillation of validation loss observed in pure CMA-ES case is mainly attributed to the nature of the adaptive algorithm.

In stage II of our proposed two-stage DFOs method, 
a relatively gentle decrease in validation loss can be observed,
demonstrating that dimension-level updates by COBYLA make the overall learning process smoother, which helps us curb the problem of fast overfitting.

\begin{figure}[H]
    \centering
    \includegraphics[scale=0.16]{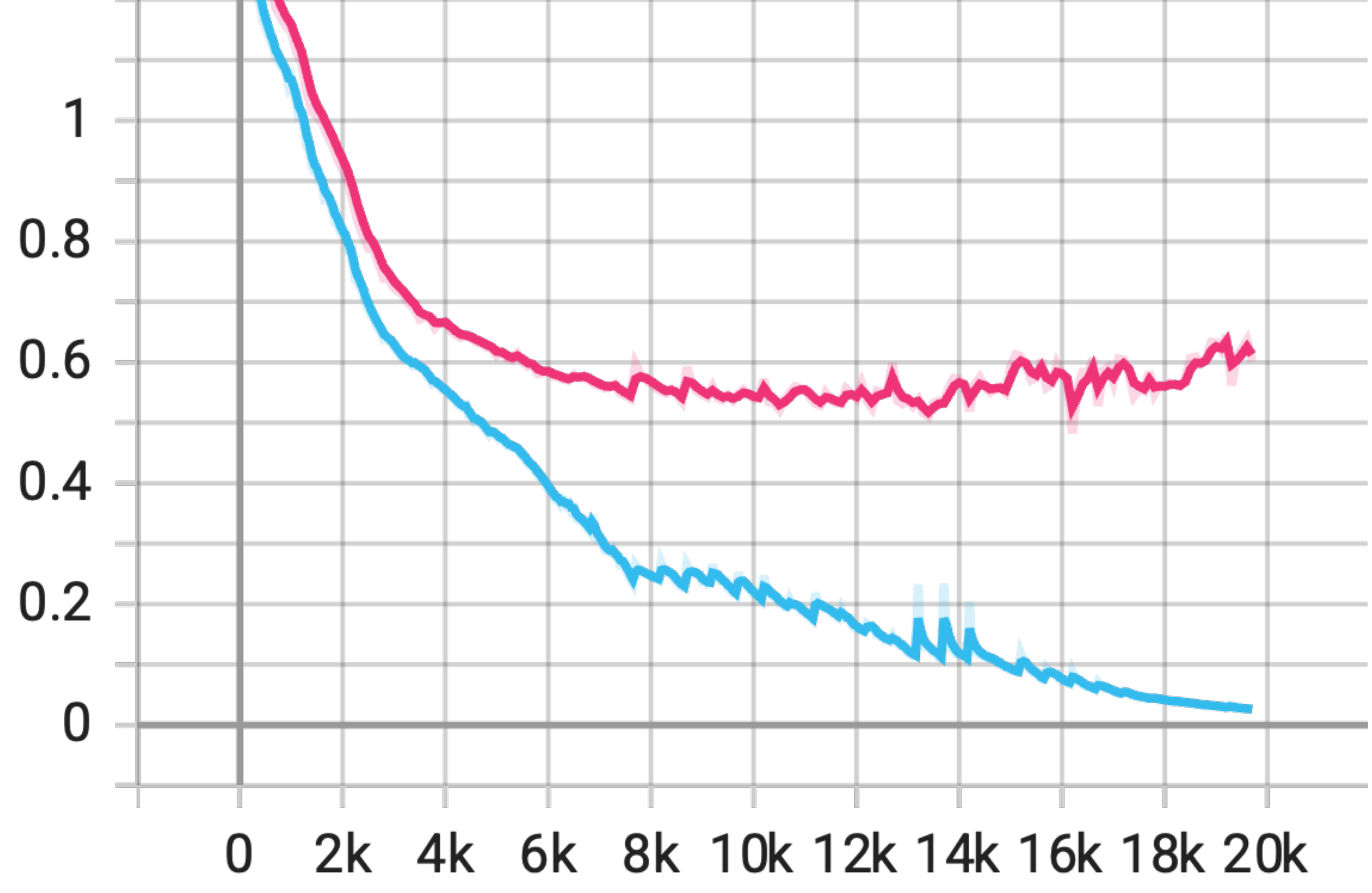}
    \caption{An illustration of Two-Stage DFOs on SST-2}
    \label{fig:sst2-2stage}
\end{figure}

Figure~\ref{fig:sst2-2stage} is another analysis of SST-2 dataset.
By employing the Two-stage DFOs (Blue line),
there is a notable improvement in the final optimization result.

\section{Experimental Details}
\label{sec:exp-details}

\subsection{Implementation}
Most of our experiments are conducted with a single NVIDIA RTX 3090 GPU. 
Due to the memory requirements, 
experiments on MRPC and DBPedia datasets are conducted with NVIDIA V100 GPUs.

\subsection{BBT-RGB Settings}
\label{sec:rgb-setting}

The details of using BBT-RGB across seven NLP tasks are listed in Table~\ref{table:rgb-settings}.
For each task, 
we report the average performance and standard deviation across three random seeds (42, 50, 66).

\begin{table*}[t]
\begin{center}
		\begin{tabular}{lccccccc}
			\toprule
			& SST-2 & Yelp & AGNews & DBPedia & SNLI & RTE &MRPC \\
			\midrule
                Two-Stage DFO & \color{ForestGreen}{\Checkmark} & \color{ForestGreen}{\Checkmark} & \color{ForestGreen}{\Checkmark} &
                \color{ForestGreen}{\Checkmark} &\color{ForestGreen}{\Checkmark} & \color{ForestGreen}{\Checkmark} & \color{BrickRed}{\XSolidBrush} \\
			M\textsuperscript{2} Verbalizers & \color{BrickRed}{\XSolidBrush} & \color{ForestGreen}{\Checkmark} & \color{BrickRed}{\XSolidBrush} & \color{ForestGreen}{\Checkmark} & \color{BrickRed}{\XSolidBrush} & \color{ForestGreen}{\Checkmark} &
			\color{ForestGreen}{\Checkmark}\\
			In\textsuperscript{2} Initialization &  \color{ForestGreen}{\Checkmark} & \color{ForestGreen}{\Checkmark} & \color{ForestGreen}{\Checkmark} & \color{ForestGreen}{\Checkmark} & \color{ForestGreen}{\Checkmark} &
			\color{ForestGreen}{\Checkmark} & \color{BrickRed}{\XSolidBrush} \\
			\bottomrule
		\end{tabular}
		\caption{The details of employing BBT-RGB, {\color{ForestGreen}{\Checkmark}} refers to use the given technique on this task, and {\color{BrickRed}{\XSolidBrush}} vice versa}
		\label{table:rgb-settings}
\end{center}
\end{table*}

\subsection{Hyperparameters Settings}
\label{sec:hyper-params}

The experimental settings in our paper are listed in Table~\ref{table:hparams}. Sigma1 and Sigma2 are the hyperparameters for CMA-ES. Alpha refers to a constant scalar for stretching the distribution of random projection matrices, as is shown in equation~\ref{eq:sigma}.

\begin{equation}
    \sigma_A = \frac{\alpha\hat{\sigma}}{\sqrt{d}\sigma_z},
    \label{eq:sigma}
\end{equation}

$\hat{\sigma}$ is the standard deviation of word embeddings from RoBERTa-Large, and $\sigma_z$ is the standard deviation of the normal distribution maintained by the CMA-ES algorithm.
The random projection matrices are frozen during the whole optimization process.

\begin{table*}[t]
\begin{center}
\resizebox{0.8\textwidth}{!}{
		\begin{tabular}{lccccc}
			\toprule
			Task & Budget1 (CMA-ES) & Budget2 (COBYLA) & Alpha & Sigma1 & Sigma2 \\
			\midrule
            SST-2 & 7,000 & 6,000 & 0.5 & 0.7 & 0.7  \\
            Yelp & 8,000 & 6,000 & 0.9 & 0.4 & 0.2  \\
            \cdashline{1-6}
            AGNews & 8,000 & 6,000 & 0.1 & 0.6 & 0.2  \\
            DBPedia & 8,000 & 6,000 & 0.3 & 0.2 & 0.2 \\
            \cdashline{1-6}
            SNLI & 8,000 & 6,000 & 0.5 & 0.45 & 0.2  \\
            RTE & 8,000 & 6,000 & 0.5 & 1 & 0.2  \\
            MRPC & 8,000 & 0 & 0.3 & 0.3 & 0.2  \\
			\bottomrule
		\end{tabular}
		}
		\caption{Hyperparameter Settings for BBT-RGB in different tasks.}
		\label{table:hparams}
\end{center}
\end{table*}

\subsection{Templates and Verbalizers}

The templates and verbalizers we employed are listed in Table~\ref{tab:template-verb} and Table~\ref{tab:m2verbalizer} respectively.

\begin{table}[H]
\centering
\resizebox{0.8\linewidth}{!}{
\begin{tabular}{ll}
\toprule
\textbf{Dataset} & \textbf{Template}     \\ 
\midrule
SST-2  & $\langle P\rangle$ $\langle S\rangle$. It was \texttt{[MASK]}  \\
Yelp P.          & $\langle P\rangle$ $\langle S\rangle$. It was \texttt{[MASK]} \\
AGNews         & $\langle P\rangle$ \texttt{[MASK]} News: $\langle S\rangle$   \\
DBPedia & {$\langle P\rangle$ [Category: \texttt{[MASK]}] $\langle S\rangle$} \\
MRPC             & $\langle P\rangle$ $\langle S_1\rangle$ ? \texttt{[MASK]}, $\langle S_2\rangle$    \\
RTE & $\langle P\rangle$ $\langle S_1\rangle$ ? \texttt{[MASK]}, $\langle S_2\rangle$ \\
SNLI    & $\langle P\rangle$ $\langle S_1\rangle$ ? \texttt{[MASK]}, $\langle S_2\rangle$ \\
\bottomrule
\end{tabular}
}
\caption{Prompt templates used in this paper. $\langle P\rangle$ is a sequence of continuous prompt tokens. $\langle S\rangle$ is the original input text.}
\label{tab:template-verb}
\end{table}

\begin{table*}[t]
\centering
\resizebox{0.6\linewidth}{!}{
\begin{tabular}{ll}
\toprule
\textbf{Dataset} & \textbf{M\textsuperscript{2} Verbalizers} \\ 
\midrule
\multirow{2}{*}{SST-2}  & \textbf{Positive}: exciting, all, indeed, $\cdots$   \\
& \textbf{Negative}: ridiculous, worse, stupid, $\cdots$ \\
\cdashline{1-2}
\multirow{2}{*}{Yelp P.} & \textbf{Positive}: addictive, sensational, classic, $\cdots$ \\
& \textbf{Negative}: boring, worse, ugly, $\cdots$  \\
\cdashline{1-2}
\multirow{4}{*}{AG's News} & \textbf{World}: South, China, Africa, $\cdots$ \\
& \textbf{Sports}: Athletics, SPORTS, Sporting, $\cdots$ \\
& \textbf{Business}: Banking, Manufacturing, Trade, $\cdots$ \\
& \textbf{Tech}: Digital, Internet, Tech, $\cdots$ \\
\cdashline{1-2}
\multirow{14}{*}{DBPedia} & \textbf{Company}: Business, Products, $\cdots$ \\
& \textbf{Educational/Institution}: Education, Schools, $\cdots$ \\
& \textbf{Artist}: Artists, $\cdots$ \\
& \textbf{Athlete}: Profile, $\cdots$ \\
& \textbf{Office Holder}: Politics, $\cdots$ \\
& \textbf{Mean Of Transportation}: Vehicles, $\cdots$ \\
& \textbf{Building}: Architecture, $\cdots$ \\
& \textbf{Natural Place}: Lakes, $\cdots$ \\
& \textbf{Village}: Rural, $\cdots$ \\
& \textbf{Animal}: Animals, Birds, $\cdots$ \\
& \textbf{Plant}: Plants, plants, Flowers, $\cdots$ \\
& \textbf{Album}: Album, Records, $\cdots$ \\
& \textbf{Film}: Movies, Films, $\cdots$ \\
& \textbf{Written Work}: Books, Fiction, $\cdots$ \\
\cdashline{1-2}
\multirow{2}{*}{MRPC} & \textbf{Equivalent}: Finally, Notably, Next, $\cdots$ \\
& \textbf{Not Equivalent}: Instead, Although, That, $\cdots$ \\
\cdashline{1-2}
\multirow{2}{*}{RTE} &  \textbf{Yes}: Indeed, So, Wordwide, $\cdots$ \\
& \textbf{No}: Also, Now, meanwhile, $\cdots$ \\
\cdashline{1-2}
\multirow{3}{*}{SNLI} & \textbf{Yes}: Whatever, YES, Regardless, $\cdots$ \\
& \textbf{Maybe}: Imagine, Usually, Typically, $\cdots$ \\
& \textbf{No}: Besides, Unfortunately, Surprisingly, $\cdots$ \\
\bottomrule
\end{tabular}
}
\caption{Examples of the M\textsuperscript{2} Verbalizers used in practice.}
\label{tab:m2verbalizer}
\end{table*}


\end{document}